# Learning Everywhere: A Taxonomy for the Integration of Machine Learning and Simulations

Geoffrey Fox
Indiana University
Bloomington, IN
gcf@indiana.edu

Shantenu Jha
Rutgers University and
Brookhaven National Laboratory
shantenu.jha@rutgers.edu

*Abstract*—We present a taxonomy of research on Machine Learning (ML) applied to enhance simulations together with a catalog of some activities. We cover eight patterns for the link of ML to the simulations or systems plus three algorithmic areas: particle dynamics, agent-based models and partial differential equations. The patterns are further divided into three action areas: Improving simulation with Configurations and Integration of Data, Learn Structure, Theory and Model for Simulation, and Learn to make Surrogates.

*Keywords—HPC, Machine Learning, Simulation*

I. INTRODUCTION

*A. Introduction*

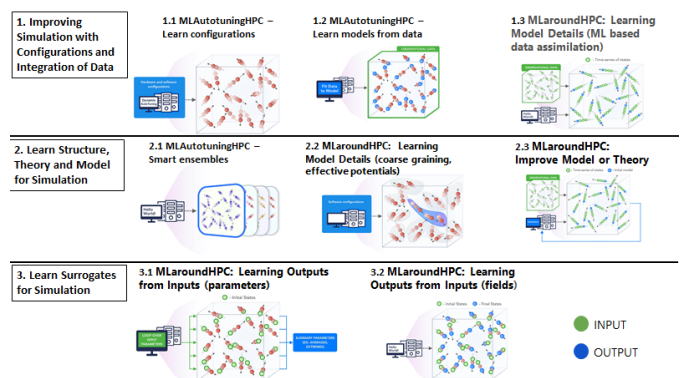

Fig. 1. The 8 MLAutotuning and MLaroundHPC scenarios described in text

This taxonomy of research at the intersection of Machine Learning and Simulations builds on papers below.
1) A quadrology of papers on learning everywhere [1]–[4]. The first paper gives an overview and the unpublished second report adds detail on Technology, Network Science, nanoengineering, biomolecular and computational biology (virtual tissues). The third paper develops the underpinnings of learning everywhere and the fourth is this paper. There are also presentations at BDEC [5] and at IPDPS [6].
2) Jeffrey Dean presentation at NeurIPS 2017 on Machine learning for systems and systems for machine learning [7]
3) Microsoft 2018 Faculty Summit presentations on AI for Systems [8], [9]
4) Satoshi Matsuoka on the convergence of AI and HPC [10]
5) An NSF funded project mainly focused on HPCforML [11], [12]

We now describe the categories used below to categorize papers [1-3], [5], [13]

- **HPCforML:** Using HPC to execute and enhance ML performance, or using HPC simulations to train ML algorithms (theory-guided machine learning), which are then used to understand experimental data or simulations.
- **MLforHPC:** Using ML to enhance HPC applications and systems

We further subdivide **HPCforML** as
- **HPCrunsML:** Using HPC to execute ML with high performance
- **SimulationTrainedML:** Using HPC simulations to train ML algorithms, which are then used to understand experimental data or simulations.

We also subdivide **MLforHPC** into several categories. First we identify
- **MLControl:** Using simulations (with HPC) in control of experiments and in objective driven computational campaigns. Here simulation surrogates of MLaroundHPC are very valuable to allow real-time predictions. This is discussed in [3]

Then can divide other aspects by whether they are before - termed MLAutotuningHPC, during the execution - termed MLaroundHPC, or after - termed MLafterHPC.
- **MLafterHPC**: ML analyzing results of HPC as in trajectory analysis and structure identification in biomolecular simulations

The other two terms where we focus in this paper are
- **MLAutotuning:** Using ML to configure (autotune) ML or HPC simulations.
- **MLaroundHPC:** Using ML to learn from simulations and produce learned surrogates for the simulations. The same ML wrapper can also learn configurations as well as results. This differs from SimulationTrainedML as the latter is typically using learned network to predict observation whereas in MLaroundHPC we are using the ML to improve the HPC performance.

Figure 1 specifies 8 subcategories in the MLAutotuning and MLaroundHPC spaces. We can take the categories in these two areas and divide them into three types of actions represented into the three rows of fig. 1 and sections 2, 3 and 4 of this detailed taxonomy paper. The three action areas are:

**II. Improving simulation with Configurations and Integration of Data**
II.A. MLAutotuningHPC – Learn configurations of system and software for particular hardware and input parameters
II.B. MLAutotuningHPC – Learn models from data at start of simulation
II.C. MLaroundHPC: Learning model details (ML based data assimilation) dynamically during simulation.

**III. Learn Structure, Theory and Model for Simulation**
III.A. MLAutotuningHPC – Smart ensembles
III.B. MLaroundHPC: Learning Model Details (coarse graining, effective potentials)
III.C. MLaroundHPC: Learning Model Details - Improving Model or Theory

**IV. Learn to make Surrogates**
Here we use ML (typically neural networks) to learn the function representing the output of the simulation.
IV.A. MLaroundHPC: Learning Outputs from Inputs (parameters)
IV.B. MLaroundHPC: Learning Outputs from Inputs (fields)

These clean atomic categories can appear differently as they are applied dynamically or differently in different (space or time) parts of simulation.

In later sections, ABM stands for Agent-Based Simulations and Data-driven Approaches to ABM systems. The work is divided into three broad application areas: Particle dynamics, ABM and Partial Differential Equation based problems. We list MLAutotuningHPC and MLaroundHPC references divided by these 3 application areas and the 8 categories summarized in Fig. 1. In this and following 8 expanded figures we use a prototypical particle dynamics simulation to represent the ML interaction with green representing input and blue output of interaction.

## II. TAXONOMY OF MLAUTOTUNING AND MLAROUNDHPC: IMPROVING SIMULATION WITH CONFIGURATIONS AND INTEGRATION OF DATA

### A. *MLAutotuningHPC – Learn configurations*

Figure 2 illustrates this category, which is classic Autotuning and one optimizes some mix of performance and quality of results with the learning network inputting the configuration parameters of the computation. The configuration includes initial values and also dynamic choices such as block sizes for cache use, variable step sizes in space and time. This category can also include discrete choices as to the type of solver to be used.

*1) Particle Dynamics-MLAutotuningHPC – Learn configurations*
1. Nanoparticle simulations using *ML* to improve performance [14]

### B. *MLAutoTuningHPC: Learning Model Setups from Observational Data*

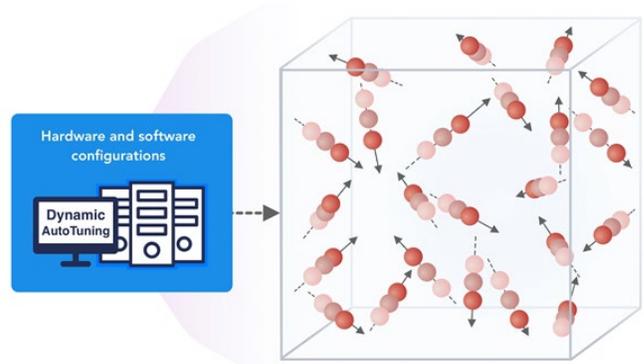

Fig. 2. MLAutotuningHPC – Learn configurations

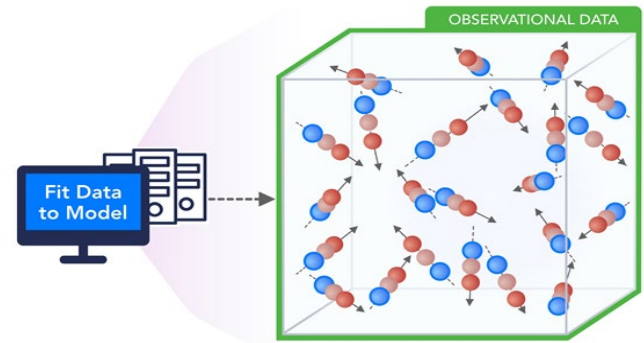

Fig. 3. MLAutoTuningHPC: Learning Model Setups from Observational Data

This category is seen when a simulation set up as a set of agents, perhaps each representing a cell in a virtual tissue simulation. Tuning agent (model) parameters to optimize agent outputs to available empirical data presents one of the greatest challenges in model construction. As well as directly setting cell parameters, one can use ML to learn the dynamics of cells replacing detailed computations by ML surrogates. As there can be millions to billions of such agents the performance gain can be huge as each agent uses the same learned model. In this case one is using MLaroundHPC: Learning Outputs from Inputs for cells or alternately MLAutotuning for multi-cell (tissue) built from the cells.

*1) Particle Dynamics-MLAutotuningHPC – Learning Model Setups from Observational Data*
2. Use of ANN's to represent dynamics of robots [15]

*2) ABM-MLAutotuningHPC – Learning Model Setups from Observational Data*
3. Machine-learning (XGBoost) and intelligent sampling to build a surrogate meta-model to calibrate agent-based models with data [16]

4. Using machine learning (modest emphasis) to represent cell (agent) behavior based on data for prediction of cancer cell behavior [17]
5. Automatic inference of a model of the escape response behavior in a roundworm directly from time series data [18] building on [19], [20]. The unknown parameters in a set of ODE's are determined by fitting data in a hierarchical fashion

*3) PDE-MLAutotuningHPC – Learning Model Setups from Observational Data*
6. Use ANN's to discover the PDE form of biological transport equations from noisy data. [21]

## C. MLaroundHPC: Learning Model Details - ML for Data Assimilation (predictor-corrector approach)

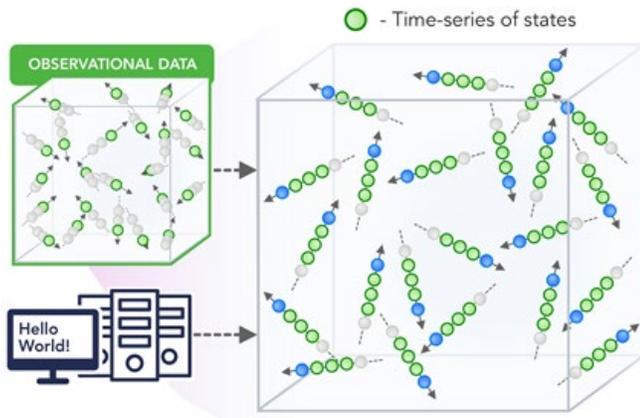

Fig. 4. MLaroundHPC: Learning Model Details - ML for Data Assimilation (predictor-corrector approach)

Data assimilation involves continuous integration of time dependent simulations with observations to correct the model with a suitable combined data plus simulation model. This is for example common practice in weather prediction field. We see this approach becoming even more important with new machine learning approaches now available and under intense research for many time series based problems such as work on ride hailing [22]. Such current state of the art expresses the spatial structure as a convolutional neural net and the time dependence as recurrent neural net (LSTM). We expect this category to grow in importance and interest. This category extends the previous one in sec. 2.1.2 with dynamic interplay between model and data.

Often the data consists of "videos" recording observational data, which is a high dimensional (spatial extent) time series. Then as a function of time one iterates a predictor corrector approach, where one time steps models and at each step optimize the parameters to minimize divergence between simulation and ground truth data. As an example considered by a team led by Glazier at Indiana University, one produces a generic agent-based model organism such as an embryo. Then one could take this generic model as a template and learn the different adjustments for particular individual organisms.

*1) ABM-MLaroundHPC: Learning Model Details (ML based data assimilation)*
7. Using data to predict solutions of complex coupled Agents for metabolic pathway dynamics [23]
8. Deep Learning RNN and CNN to predict epidemics viewed as time series [24]
9. LSTM based Flu epidemic forecasting enhanced by environmental data such as climate [25]

*2) PDE-MLaroundHPC: Learning Model Details (ML based data assimilation)*
10. Deep Learning to find sub-grid processes (such as cloud processes) for Climate prediction [26]

## III. TAXONOMY OF MLAROUNDHPC: LEARN STRUCTURE, THEORY AND MODEL FOR SIMULATION

### A. MLAutotuningHPC – Smart ensembles

Here we choose the best strategy to achieve some computation goal such as providing the most efficient training set with defining parameters spread well over the relevant phase space. Ensembles are also essential in many computational studies such as weather forecasting or search for new drugs where regions of defining parameters need to be searched. This category overlaps with the following Learning Model Details (effective potentials and coarse graining) category as both look at the structure of the simulation. Different papers tackle related but distinct goals. Some look for reaction coordinates that are collective variables (CV) that can be used to accelerate the simulation; these are typically the slowest varying with time modes of the system. Others look for structure (order parameters) of the system such as "has the protein folded".

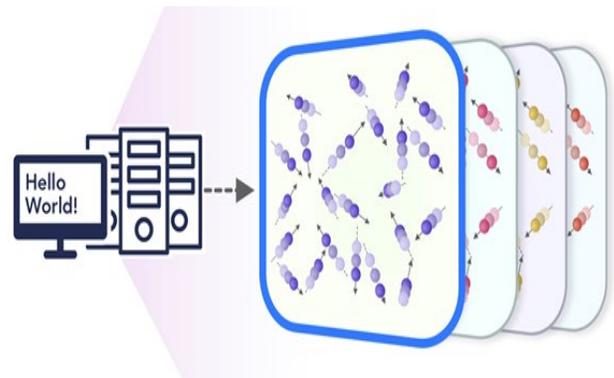

Fig. 5. *MLAutotuningHPC – Smart ensembles*

*1) General Simulations-MLAutotuningHPC – Smart ensembles*
11. Use of visualization to control smart ensembles of simulations [27]

*2) Particle Dynamics-MLAutotuningHPC – Smart ensembles*
12. Review of techniques for smart ensembles [28]

13. Use of machine learning to guide molecular dynamics simulations to explore full range of phase space [29]. Manifold learning is used to find a low dimension set of collective variables and then to learn dynamics in those variables.
14. Use of Machine Learning (Best Arm Identification method) to optimize determination of protein-ligand binding (docking) energies when total compute resources are constrained, [30]
15. Efficient exploration of configuration space by adding an adaptively computed biasing potential using machine learning to the original dynamics. [31]–[35]
16. Use of the "information Bottleneck" approach to design an ANN that will identify a collective coordinate that will guide simulations with importance sampling to correct bias [36], [37].This leads to a collective coordinator with good physical (chemical) interpretation.
17. Loop over multiple molecular dynamics and Deep Learning steps to more accurately sample phase for long time computations - termed "Reweighted autoencoded variational Bayes for enhanced sampling (RAVE)" [38], [39]
18. Use reinforcement learning to learn a ANN representation of the Free Energy based on an uncertainty estimate comping from a set of ANN's with the same updates and different random starting weights [40]. The choice of collective variables (CV) is not discussed except to note that approach can accommodate a quite large number (10-20) of CV's.
19. Study of protein folding using machine learning to identify the special regions of phase space where proteins do indeed fold [41], [42]. Google's Alphafold [43], [44] won [45] the 13th Critical Assessment of Structure Prediction (CASP) competition [46] with deep learning used to identify how specific proteins fold. Such studies can be followed up by traditional MD simulations. In [47] convolutions and a variational autoencoder (CVAE) are used for dimension reduction to identify folding region.

*3) ABM-MLAutotuningHPC – Smart ensembles*
20. Smart ensembles for cancer agent-based models with PhysiCell. [48]

B. *MLaroundHPC: Learning Model Details (effective potentials and coarse graining)*

This is classic coarse graining strategy with recently, deep learning replacing dimension reduction techniques.) One can learn effective potentials and interaction graphs. An effective potential is an analytic, quasi-empirical or quasi-phenomenological potential that combines multiple, perhaps opposing, effects into a single potential.

*1) Particle Dynamics-MLaroundHPC: Learning Model Details (effective potentials)*
21. Use of machine learning to generate an effective Hamiltonian using initial local updates as training data to choose correlated update spins with Wolff's method near a critical point [49]. This is applied in [50]
22. Neural-network representation [51]–[54] of DFT potential-energy surfaces
23. General framework for calculating a many-body coarse-grained potential. [55]
24. Formulate coarse-graining as a supervised machine learning problem and use coarse-graining error and cross-validation to select and compare the performance of different models. [56]
25. Review of the use of neural networks to represent potentials and speed up simulations [57]. Has plot of physics, chemistry and materials papers per year using ANN's. There are 1500 per year after 2010.

*2) Particle Dynamics-MLaroundHPC: Learning Model Details (coarse graining)*
26. VAMP(variational approach for Markov processes)nets to learn end to end reduced complexity surrogates of molecular dynamics without custom modelling such as transformation of simulated coordinates into structural features, dimension reduction, clustering the dimension-reduced data, and estimation of a Markov state models [58]
27. Use of collective variables (dimension reduction) to study protein dynamics [59]
28. Obtains one-dimensional collective variables for studying rarely occurring transitions between two metastable states separated by a high free energy barrier [60].
29. Collective variables to sample molecular dynamics and free energy landscape using autoencoders [61]–[64]. Includes MLAutotuningHPC – Smart ensembles
30. Use of machine learning to support long time scale molecular simulations [65] Reviews other approaches such as RAVE and VAMP. Includes MLAutotuningHPC – Smart ensembles.

*3) PDE-MLaroundHPC: Learning Model Details (coarse graining)*
31. Use of equation free modeling [66] for coarse graining combined with manifold learning (dimension reduction) [67]
32. Uses neural nets as expansion functions for solutions of partial differential equations [68].

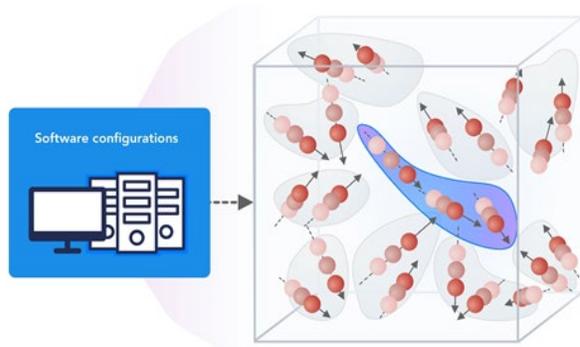

Fig. 6. MLaroundHPC: Learning Model Details (effective potentials and coarse graining)

*C. MLaroundHPC: Learning Model Details - Inference of Missing Model Structure*

The final category in the Structure, Theory and Model class and represented in the above figure imagines a future where AI will essentially be able to derive theories from data, or more practically a mix of data and models. This is especially promising in agent based models which often contain phenomenological approaches such as the predictor-corrector method of sec. 2.1.3. We expect that will take the results of such assimilation and effective potentials and interactions discussed earlier and use them as the master model or theory for future research.

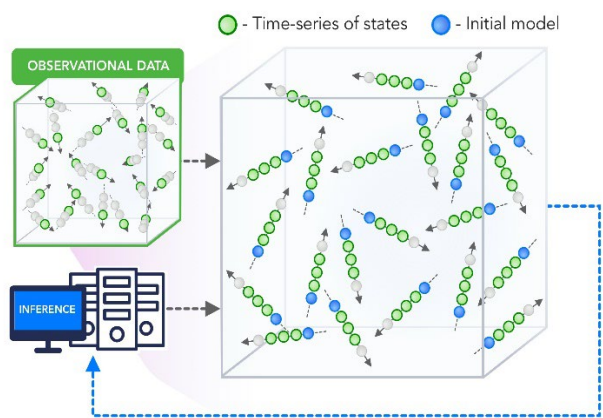

Fig. 7. MLaroundHPC: Learning Model Details - Inference of Missing Model Structure

IV. TAXONOMY OF MLAUTOTUNING AND MLAROUNDHPC: LEARN SURROGATES FOR SIMULATION

*A. MLaroundHPC: Learning Outputs from Inputs: a) Computation Results from Computation defining Parameters*

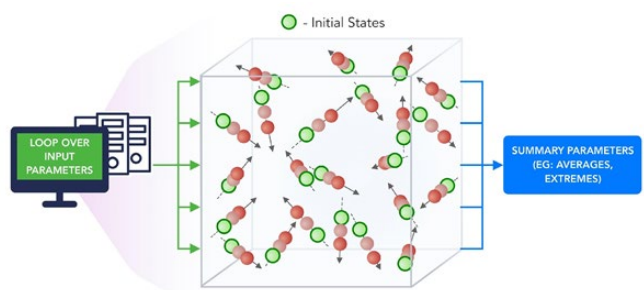

Fig. 8. MLaroundHPC: Learning Outputs from Inputs: Computation Results from Computation defining Parameters

In this category, one just feeds in a modest number of meta-parameters that define the problem and learn a modest number of calculated answers. In many circumstances, summary parameters are joined with observed properties to specify compounds. This task presumably requires fewer training samples than "fields from fields" (next category) and is main MLaroundHPC use so far.

Operationally this category is the same as SimulationTrainedML but with a different goal: In SimulationTrainedML the simulations are performed to directly train an AI system rather than the case here where the AI system is being added to learn a simulation.

*1) Particle Dynamics- MLaroundHPC: Learning Outputs from Inputs (parameters)*

33. An early paper in 2012 using non-ANN machine learning to learn energies from molecular properties [69]
34. Use of generative and predictive ANN to predict drug properties from their SMILES representation using existing databases [70].Use of DNN to learn crystal energies and stability with training data calculated by DFT. [71]
35. Review of machine learning (emphasized) for molecular and materials science [72]
36. Nanoparticle simulations [73] defining surrogates learnt as a function of defining parameters
37. Review article on machine learning to predict material properties from structure of compounds. Uses observation and simulations to determine structure-property relationships for training [74]
38. Use of neural nets to describe potentials and simulation results for Infrared Spectra [75] The input features to the ANN's are the parameters of Frenkel exciton Hamiltonians and the output average exciton transfer times and overall transfer efficiencies.
39. Machine Learning (kernel ridge regression) to map database (of DFT simulations) into material properties. [76]
40. Machine Learning (kernel ridge regression) to map database (of DFT simulations) into valence charge densities. [77], [78]
41. ANN's for fast estimate of excitation energy transfer properties (used in solar cells) [79]. The ANN is used to map Hamiltonian specifications into material properties.
42. Machine Learning to predict the energies and forces and avoid repetitive computations [80]. A decision engine decides whether to use learnt result or calculate using full simulation.
43. Machine Learning used to estimate forces in molecular simulations choosing between ab initio Quantum mechanics or regression based ML estimate from a database enhanced dynamically. [32], [81]
44. Review of machine learning with dimensionality reduction and clustering algorithms, drug discovery DeepTox, free-energy surface of molecules, ligand binding site detection, ligand pose prediction, ligand, active/inactive classification, ligand binding affinity prediction, and protein design, DeepChem software, MoleculeNet challenge and access to relevant QSAR prediction datasets. Two cases covered in detail - ML representation of Quantum forces and prediction of binding affinities. [82]
45. Deep Learning to study compositional and configurational chemical space for molecules of

intermediate size. Focus on use of a particular representation of input molecular structure [83].
46. Specifying atom representations for input into machine learning [84]
47. *MLaroundHPC: Learning Outputs from Inputs (parameters)* is reviewed but generalized to learn system wavefunction in its hamiltonian matrix element form allowing richer set of predictions with *MLaroundHPC: Learning Outputs from Inputs (fields)* [85]

*2) PDE-MLaroundHPC - Learning Outputs from Inputs (parameters)*

48. Finding coefficients of a PDE that reproduce observed data [86]
49. Machine Learning surrogates of heart simulations to speed up aortic aneurysm studies [87]

B. *MLaroundHPC: Learning Outputs from Inputs: b) Fields from Fields*

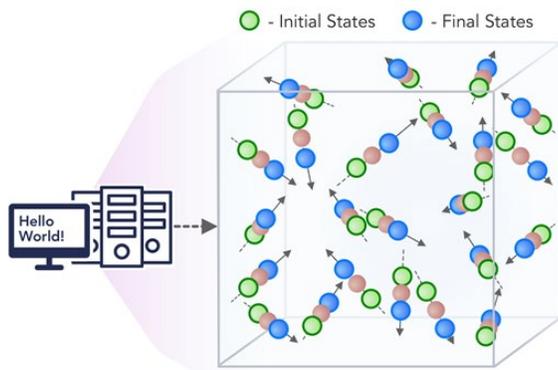

Fig. 9. MLaroundHPC: b) Learning Outputs from Inputs: Fields from Fields

Here one feeds in initial conditions and the neural network learns the result where initial and final results are fields

There is also a mixed category c) Learning Outputs from Inputs: output fields from computation defining parameters combining a) and b), which we don't illustrate.

*1) Particle Dynamics-MLaroundHPC: Learning Outputs from Inputs (fields)*

50. An early paper using in 1994 neural nets to solve ODE's. [88]]
51. Investigation of different neural network structures to learn the results of an Ising model simulation near its critical point comparing with classic Monte Carlo using a combination of single-site Metropolis and Wolff cluster updates [89]
52. MLaroundHPC: Learning Outputs from Inputs (parameters) is reviewed but generalized to learn system wavefunction in its hamiltonian matrix element form allowing richer set of predictions with MLaroundHPC: Learning Outputs from Inputs (fields) [85]
53. Using Generative Adversarial Networks to produce surrogates of large scale simulations of the effect of gravitational lensing used to study early universe CosmoGAN [90], [91] with supplement [92] on Github

54. Uses LSTM's to learn time series represented by molecular dynamics simulation [93]. Promising results on small model systems.
55. Uses deep learning to find a clean set of collective coordinates that can be easily sampled to efficiently move through phase space [94].

*2) ABM-MLaroundHPC: Learning Outputs from Inputs (fields)*

56. Use of Deep Learning LSTM to produce surrogates of a one-dimensional biological agent simulation [95]. Errors were estimated by training four neural networks differing in initial (random) choices of weights. 105 simulations took 2 months on a 400 node cluster and were followed by looking at 108 surrogate runs for an in depth survey over the full phase space. The speedup was 30,000 using surrogates.
57. Deep Learning for Agent-based Epidemic Forecasting DEFSI with ANN's learning detailed (county level) information from simulations. [96]

*3) PDE-MLaroundHPC: Learning Outputs from Inputs (fields)*

58. Finding forward (direct) and inverse mapping functions of input to output. The inverse map is particularly interesting as it is no harder than direct method for ANN's but classic PDE solvers only give direct map straightforwardly. [97], [98]
59. Deep Learning for solving partial differential equations [99], [100] (called Physics Informed Neural Net PINN) extended to nonlinear systems [101]
60. Uses PINN to solve stochastic forward and inverse problems with separate DNN to learn error. [102]
61. Deep learning to find surrogates for fluid flow simulations [103]
62. Use of machine learning to improve Extended dynamic mode decomposition for representing Koopman Operator to represent dynamical systems. The ANN learns the operators used to represent the solution.[104]
63. Solving high dimensional (up to 1000's) partial differential equations using deep learning surrogates with differentiation of neural net form and no mesh points. Exact solutions used to train surrogates [105], [106]
64. Explicitly differentiating the ANN in [87] solving advection and diffusion type PDEs in complex geometries[107]

V. CONCLUSIONS

We have reviewed 107 references in 64 distinct micro categories. These are grouped into 8 action areas and separately discussed for particles, agent-based modelling and partial differential equation solvers. We see that deep learning is showing striking success and tends to replace other machine learning approaches. As discussed in [3], we expect these successes to lead to large increases in effective performance (often by several orders of magnitude) and lead computational science to new discoveries. Although we have broken out methods into the 8 categories, one can expect them all to be combined in future projects as illustrated in fig. 10.

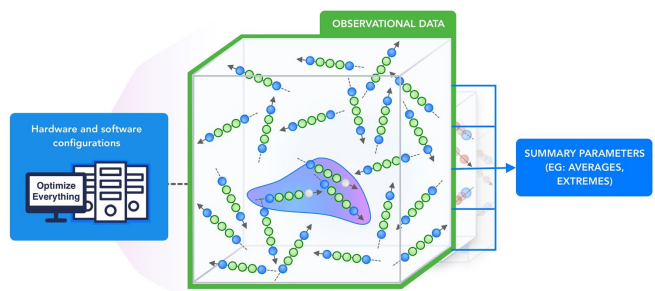

Fig. 10. 8 MLAutotuning and MLaroundHPC appoaches combined.

**Acknowledgements**
Partial support by NSF CIF21 DIBBS 1443054, NSF nanoBIO 1720625, NSF CINES 1835598 and NSF BDEC2 1849625 is gratefully acknowledged. We thank the "Learning Everywhere" collaboration James A. Glazier, JCS Kadupitiya, Vikram Jadhao, Minje Kim, Judy Qiu, James P. Sluka, Endre Somogyi, Madhav Marathe, Abhijin Adiga, Jiangzhuo Chen, and Oliver Beckstein for many discussions. SJ is partially supported by DOE ECP "ExaLearn".